\documentclass[conference]{IEEEtran}
\IEEEoverridecommandlockouts
\usepackage{cite}
\usepackage{amsmath,amssymb,amsfonts}
\usepackage{algorithmic}
\usepackage{graphicx}
\usepackage{textcomp}
\usepackage{xcolor}
\def\BibTeX{{\rm B\kern-.05em{\sc i\kern-.025em b}\kern-.08em
    T\kern-.1667em\lower.7ex\hbox{E}\kern-.125emX}}
\begin{document}

\title{Faster, Lighter, More Accurate:\\A Deep Learning Ensemble for Content Moderation}

\author{
\IEEEauthorblockN{Mohammad (Arian) Hosseini}
\IEEEauthorblockA{\textit{Comcast, University of Illinois at Urbana-Champaign} \\
USA \\
shossen2@illinois.edu}
\and
\IEEEauthorblockN{Mahmudul Hasan}
\IEEEauthorblockA{\textit{Comcast} \\
USA \\
mhasan@comcast.com
}
}

\maketitle

\begin{abstract}
To address the increasing need for efficient and accurate content moderation, we propose an efficient and lightweight deep classification ensemble structure. Our approach is based on a combination of simple visual features, designed for high-accuracy classification of violent content with low false positives. Our ensemble architecture utilizes a set of lightweight models with narrowed-down color features, and we apply it to both images and videos.

We evaluated our approach using a large dataset of explosion and blast contents and compared its performance to popular deep learning models such as ResNet-50. Our evaluation results demonstrate significant improvements in prediction accuracy, while benefiting from 7.64x faster inference and lower computation cost.

While our approach is tailored to explosion detection, it can be applied to other similar content moderation and violence detection use cases as well. Based on our experiments, we propose a \textit{"think small, think many"} philosophy in classification scenarios. We argue that transforming a single, large, monolithic deep model into a verification-based step model ensemble of multiple small, simple, and lightweight models with narrowed-down visual features can possibly lead to predictions with higher accuracy.
\end{abstract}


\section{Introduction}
\label{Introduction}

Automated content moderation has become an essential aspect of online platforms in recent years, with explosive growth in user-generated content. As video-sharing websites and online marketplaces have become popular, they have also become a hub for the dissemination of videos containing explosions, which could be disturbing and harmful to younger audiences. Content moderation, therefore, is crucial to ensure user safety and compliance with laws and regulations, particularly for video broadcasting where scenes of explosions are often depicted. One important aspect of content moderation in this context is the detection of unsafe scenes for kids, which includes identifying explosive content and other forms of violent or disturbing imagery. Automated explosion detection techniques are crucial for enabling quick and effective responses to ensure public safety and security, and to prevent the spread of harmful content on online platforms.
\begin{figure}[!t]
\centering
\includegraphics[width=1\columnwidth]{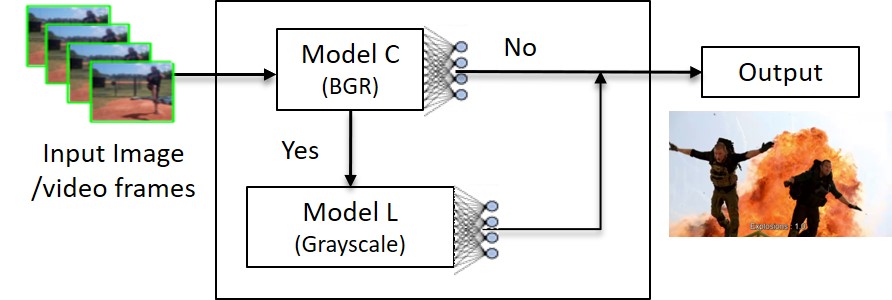}
%\includegraphics[width=.3\textwidth]{explosion.png}
%\vspace{-0.5cm}
\caption{An abstract structure of our verification-based ensemble model.}
\label{fig:abstract}
%\vspace{-0.5cm}
\end{figure}

In this context, there has been a significant interest in applying deep neural networks to address many important real-world computer vision problems, including automated content moderation and explosion detection. However, to achieve higher accuracy, the complexity of neural networks has increased dramatically over time, leading to a challenge in terms of memory and computational resources. Moreover, larger datasets are required to provide higher prediction accuracy and address false positives. While ensemble learning techniques are designed to improve classification accuracy, they often create a subset dataset from the original dataset, leading to the same results since they are getting the same input. Therefore, ensemble models may not perform well without feature engineering. 

In this paper, we propose a novel ensemble structure consisting of two lightweight deep models that are designed to perform high-accuracy and fast classification for both image and video classification use cases, specifically targeting explosion detection. Our design uses a verification-based combination of two lightweight deep models, each individual model making predictions on a different color feature: the main color-based model which operates based on 3 RGB color channels, and a secondary structure-oriented model which operates based on a single grayscale channel, focusing more on the shape of the object through intensity than learning about their dominant colors. We implemented and evaluated our approach for explosion detection use case where video scenes containing explosions are identified. Our evaluation results based on experiments on a large test set show considerable improvements in classification accuracy over using a ResNet-50 model, while benefiting from the structural simplicity and significant reduction in inference time as well as the computation cost by a factor of 7.64.

\begin{figure}[!b]
\centering
\includegraphics[width=0.48\columnwidth]{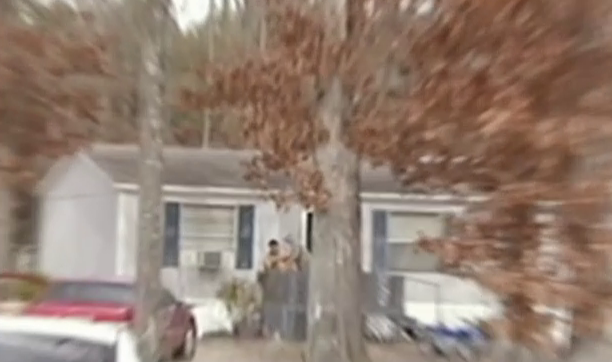}~\includegraphics[width=0.48\columnwidth]{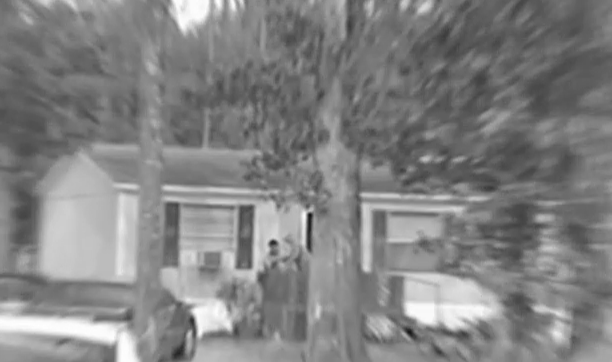}
\caption{An example false detection when making prediction using a grayscale model. When in grayscale, the image structure resembles an explosion.}
%\vspace{-0.1cm}
\label{fig:falsepositive}
\end{figure}
While our approach is applied to explosion detection scenarios, it can be generalized to other similar image and video classification use cases. Based on the insights gained from our evaluations, we further make an argument to "think small, think many" which aims to beat the complexity of large models with the simplicity of smaller ones. Our approach replaces a single, large, monolithic deep model with a verification-based hierarchy of multiple simple, small, and lightweight models with step-wise contracting color spaces, possibly resulting in more accurate predictions. In this paper, we provide a detailed explanation of our approach and its evaluation results, which we believe will be beneficial for researchers and practitioners working in the field of computer vision. As automated content moderation and explosion detection are crucial for user safety and platform compliance with regulations, our proposed approach can play a vital role in making online spaces more secure for everyone. Furthermore, the increased efficiency in automated compliance and moderation can reduce the burden on manual efforts, allowing human moderators to focus on more nuanced content issues.

\begin{figure*}[!t]
\centering
%\vspace{-0.2cm}
\includegraphics[width=1\textwidth]{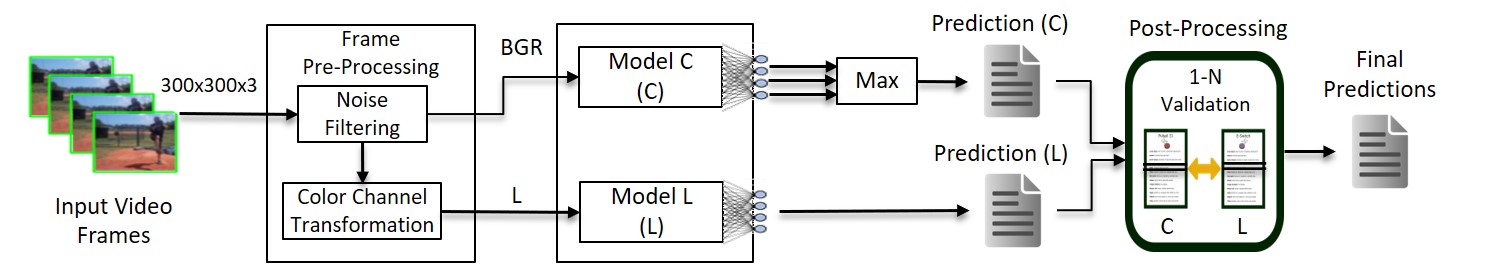}
%\vspace{-0.5cm}
\caption{Details of our proposed ensemble extended towards video analytics. C and L represent colored and gray-scale features.}
\label{fig:details}
%\vspace{-0.3cm}
\end{figure*}

\begin{figure*}[!t]
\centering
\includegraphics[width=1.0\textwidth]{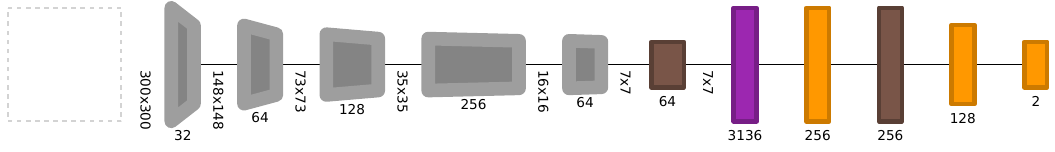}
%\vspace{-0.3cm}
\includegraphics[width=1.0\textwidth]{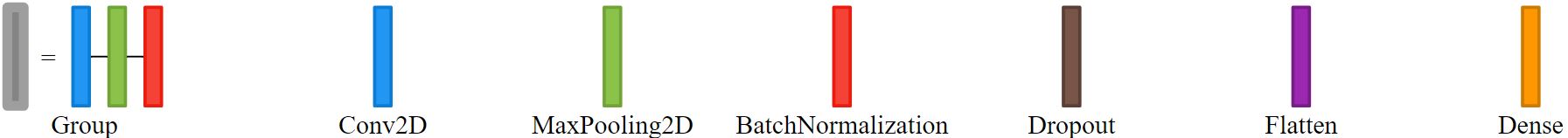}
%\vspace{0.2cm}
\caption{The internal architecture of our base models. Model C and L both use the same architecture.}
\label{fig:model}
%\vspace{-0.4cm}
\end{figure*}

The remainder of the paper is structured as follows. Section II provides an overview of relevant background information and related work. In Section III, we describe the design and architecture of our proposed ensemble approach. Section IV outlines our experimental setup and presents our evaluation results. In Section V, we discuss our findings and explain how our approach can be extended to other content moderation use cases. Finally, in Section VI, we summarize our contributions and offer concluding remarks.

\section{Related Work}
Ensemble models are generally aimed to improve the classification accuracy. Multiple models are used to make predictions for each data point, with each sub-model trained individually through variations in the input data. The predictions by each model are considered as a vote, where all votes can later be fused as a single, unified prediction and classification decision.

To achieve this, various techniques exist to combine predictions from multiple models. Voting, averaging, bagging and boosting are among the widely-used ensemble techniques \cite{ensemble1, ensemble2, ju2018relative}. In max-voting for instance, each individual base model makes a prediction and votes for each sample. Only the sample class with the highest votes is included in the final predictive class. In the averaging approaches, the average predictions from individual models are calculated for each sample. In bagging techniques, the variance of the prediction model is decreased by random sampling and generating additional data in the training phase. In boosting, subsets of the original dataset are used to train multiple models which are then combined together in a specific way to boost the prediction. Unlike bagging, here the subset is not generated randomly. While effective, a requirement in many of these broad approaches is to create a subset dataset from the original dataset, which is used to make predictions on the whole dataset. So there is a high chance that these models will give the same result since they are getting the same input.  

A body of work have shown the potential of ensemble methods in improving the performance and accuracy of deep learning models for a variety of classification tasks. In \cite{tao2019deep}, the authors propose an ensemble algorithm to address overfitting through looking at the paths between clusters existing in the hidden spaces of neural networks. The authors in \cite{wasay2018mothernets} present MotherNets, which enable training of large and diverse neural network aimed to reduce the number of epochs needed to train an ensemble. In \cite{bonab2017less}, the authors propose a geometric framework for a priori determining the ensemble size through studying the effect of ensemble size with majority voting and optimal weighted voting aggregation rules. Another approach that has been explored in the literature is the use of ensembles of weak classifiers, such as decision trees, to improve classification performance. For example, in \cite{GHIASI2021104089, Mohammed2021}, the authors investigated the decision tree-based ensemble learning in the classification of various cancers. 

In this paper, we argue that transforming a single, large, monolithic deep model into an ensemble of multiple smaller models can potentially enable higher accuracy, while benefiting from reduced training costs and faster inference time. Our proposed ensemble approach differs from these existing methods in several ways. First, we focus on a specific use case of content moderation, specifically the detection of violent or explosive content. Second, we use a set of lightweight models with narrowed-down color features, which reduces the computation cost and enables faster inference compared to larger, more complex models. Third, our ensemble architecture is designed to handle both images and videos, which is an important consideration for real-world content moderation applications. Our approach is independent, and can further be combined with other ensemble techniques discussed above.

\section{Methodology}
Our proposed methodology is designed to address the limitations of using a single model approach for image classification tasks, specifically for explosion detection. Our experimental results showed that using only a color-oriented model, which focuses on RGB color features, can result in false positives due to the misclassification of scenes with light-emitting sources. On the other hand, while a grayscale model can eliminate color-induced false positives, it introduces other false positives that are similar in structural shape to explosions or fires. To overcome these limitations, we propose a verification-based ensemble structure that combines the strengths of both models, Model C and Model L. Model C is used as the primary classifier, and its predictions are verified or validated by Model L, which is more structure-oriented and uses grayscale (L) features. This verification step helps to filter out false positives and improve overall prediction accuracy. Figure \ref{fig:abstract} illustrates the abstract design of our proposed verification-based ensemble structure, which involves step-wise contracting color spaces. In this structure, if Model C predicts an input sample image as negative, the overall prediction would result as negative, and only if Model C predicted an input sample as positive, the prediction will be verified or validated with Model L. Given these facts, our proposed architecture is designed based on two-fold insights:

Firstly, for many use cases such as those seen in our explosion image classification use case, we experienced that while the use of a single model operating on RGB-based color features provides a higher accuracy, it can wrongly classify scenes such as sunlight, lamps, or light-emitting sources as explosions. Similarly, while employing an individual model based on grayscale features can eliminate such color-induced false positives, it further introduces other false positives which are similar to explosions or fires in structural shape. Through our experimentation, we realized that a grayscale model can potentially identify bushes or trees, clay grounds, clouds, or steam as fire or explosion simply due to lack of knowledge on the color data. This problem is exacerbated in video frames due to motion blur.

Therefore, we experimented with combining both color and grayscale intensity (which focuses more on the structure) to provide a higher classification accuracy. Figure \ref{fig:falsepositive} depicts an example false positive identified as an explosion when using only the grayscale model.

Secondly, limited features, in this case, a limited number of color spaces through removing chrominance and keeping only luminance as the color feature, would generally lead to lower learnability due to a lower number of features to train on. This means the model would incur higher recall and lower precision. Our evaluation showed that passing predictions from a supposedly higher-precision model to a model with higher-recall can potentially lead to filtering out false positives and therefore increase overall prediction accuracy.

Figure \ref{fig:details} illustrates the structural details of an extended version of our ensemble design, which is applied to video frames. To process video frames, each frame is captured and resized to 300x300, and the pixels along with their 3 color channels are forwarded to a frame pre-processing phase. We chose the input dimension of 300x300 based on a trade-off between computation cost and prediction accuracy. To reduce noise, Anti-Aliasing technique is applied on every frame as a part of the pre-processing phase.

To extract features, color channels are separated from each frame, producing RGB color features (signified as C). A Color Channel Transformation module transforms the 3 RGB channels to grayscale-only features (signified as L). The original RGB features are passed to Model C directly, while the grayscale features are passed to Model L. Each model produces a binary prediction output, signifying positive (i.e., explosion in our use case) or negative (i.e., non-explosion).
After conducting a thorough evaluation of precision and recall trade-offs, we selected a prediction threshold of 90\% for both models. This threshold indicates that any detection with a score above 90\% will be considered a positive detection.

After the predictions from the individual models are made, our proposed post-processing technique applies a validation-based mechanism to the results from Model C (predictions C) and Model L (predictions L). This mechanism involves \textit{verifying} the positive predictions made by Model C by comparing them with the predictions made by Model L. Only if Model L also made a positive prediction, the overall prediction is considered positive. This step helps to filter out false positives and improve the overall accuracy of the ensemble.

After the predictions from the individual models are made, we employ a temporal coherence check to further improve the prediction accuracy for video frames. The temporal coherence check ensures that the predicted labels for subsequent frames in a video sequence are consistent. Specifically, for Model C predictions, we compute the majority vote of the labels for a set of three consecutive frames. If two or more frames in the set are predicted as positive, the majority label is considered positive. This majority value is then fused with the prediction from Model L through the same validation-based approach to derive the final outcome using a 1-N validation approach. For every positive frame $i$ (labeled as "explosion") in Model C predictions, we check the three neighbor frames $i-1$, $i$, and $i+1$  in Model L predictions. If at least one of these frames is predicted as positive (labeled as "explosion"), the final prediction is positive. While our video-specific post-processing approach can be generalized to other numbers of neighbors, such as 1 or 5, We found that our choice of three consecutive frames provides an efficient trade-off between the final precision and recall values.

Figure \ref{fig:model} depicts the internal architecture of each of the two models. Both Model C and Model L are feed-forward convolutional neural networks that consist of multiple groups of a 2D convolution layer (Conv2D), followed by a Max-Pooling layer and a Batch Normalization layer. The models have five layers with the number of convolution filters and kernel sizes as illustrated in Figure \ref{fig:model}.
The first Conv2D layer has 32 filters with a kernel size of 5x5, 64 filters and a kernel size of 3x3 for the second one, 128 filters, 256, and lastly 64 filter with kernel size of 3x3 as well for the other Conv2D layers. 

To standardize the inputs passed to the next layer and accelerate the training process while reducing the generalization error, we apply batch normalization. We also use a dropout mechanism with a rate of 0.2 to prevent possible overfitting. After passing through a Flatten layer, output features are flattened into a 1-dimensional array for entering the next layer. The produced data is then passed through three dense layers and another dropout layer to achieve the final binary prediction. Rectified Linear Unit (ReLU) is used as the activation function for all the Conv2D and Dense layers. While we designed this specific sequential neural network as our base model, we acknowledge that other lightweight models such as MobileNetV2 \cite{mobilenetv2}, SqueezeNet \cite{squeezenet}, or ShuffleNet \cite{shufflenet} can also serve as a base model for our ensemble design.

\begin{figure}[!t]
\centering
%\vspace{-0.3cm}
\includegraphics[trim=.9in 3.8in .9in 3.6in, width=0.95\columnwidth]{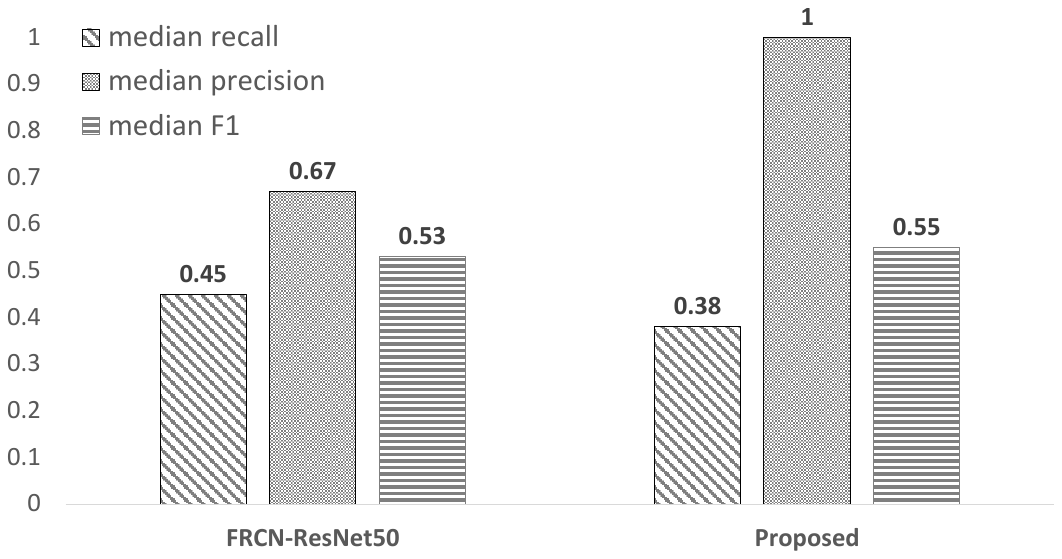}
\caption{Comparison of our proposed approach against the ResNet-50 model as used as the back-end of a Faster R-CNN detection network.}
\label{fig:evaluation}
\end{figure}

\begin{figure}[!t]
\centering
\includegraphics[trim=.75in 2.6in .75in 2.8in, width=.487\columnwidth]{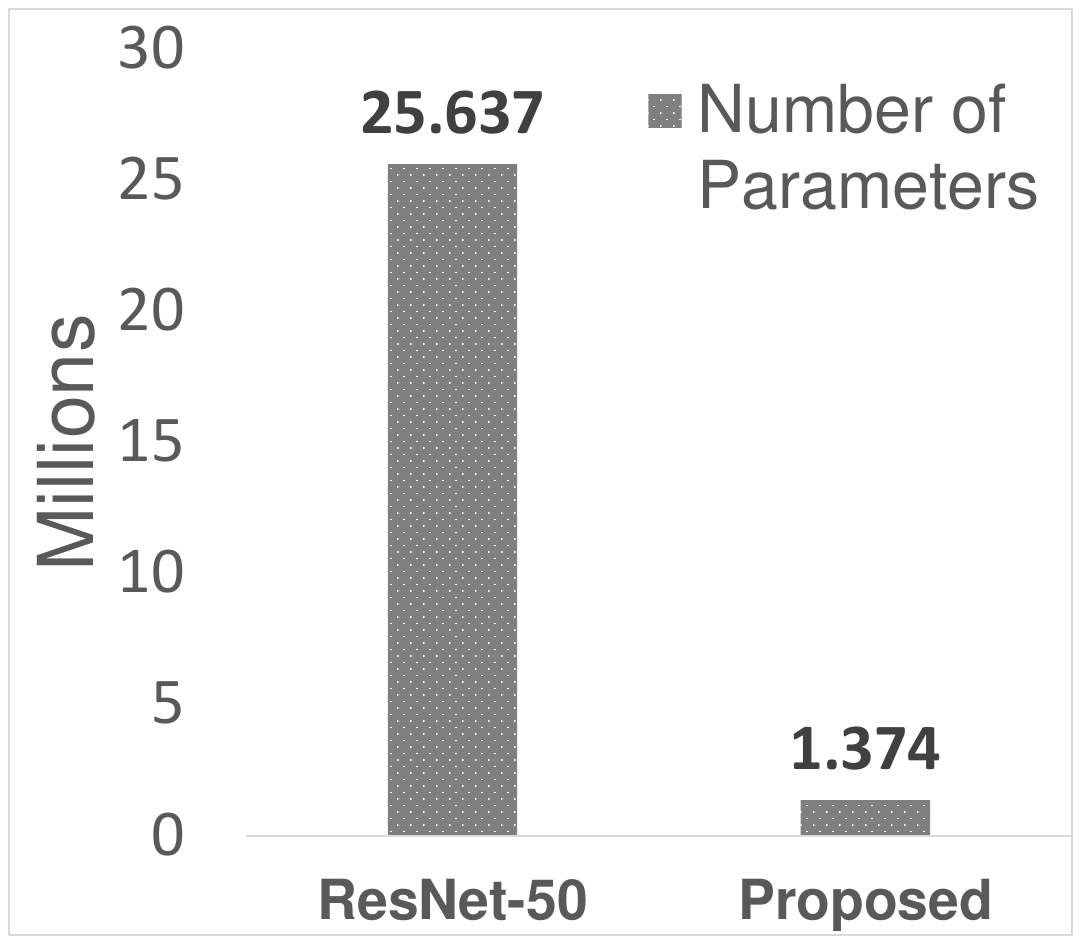}
\includegraphics[trim=.75in 2.6in .75in 2.8in, width=.49\columnwidth]{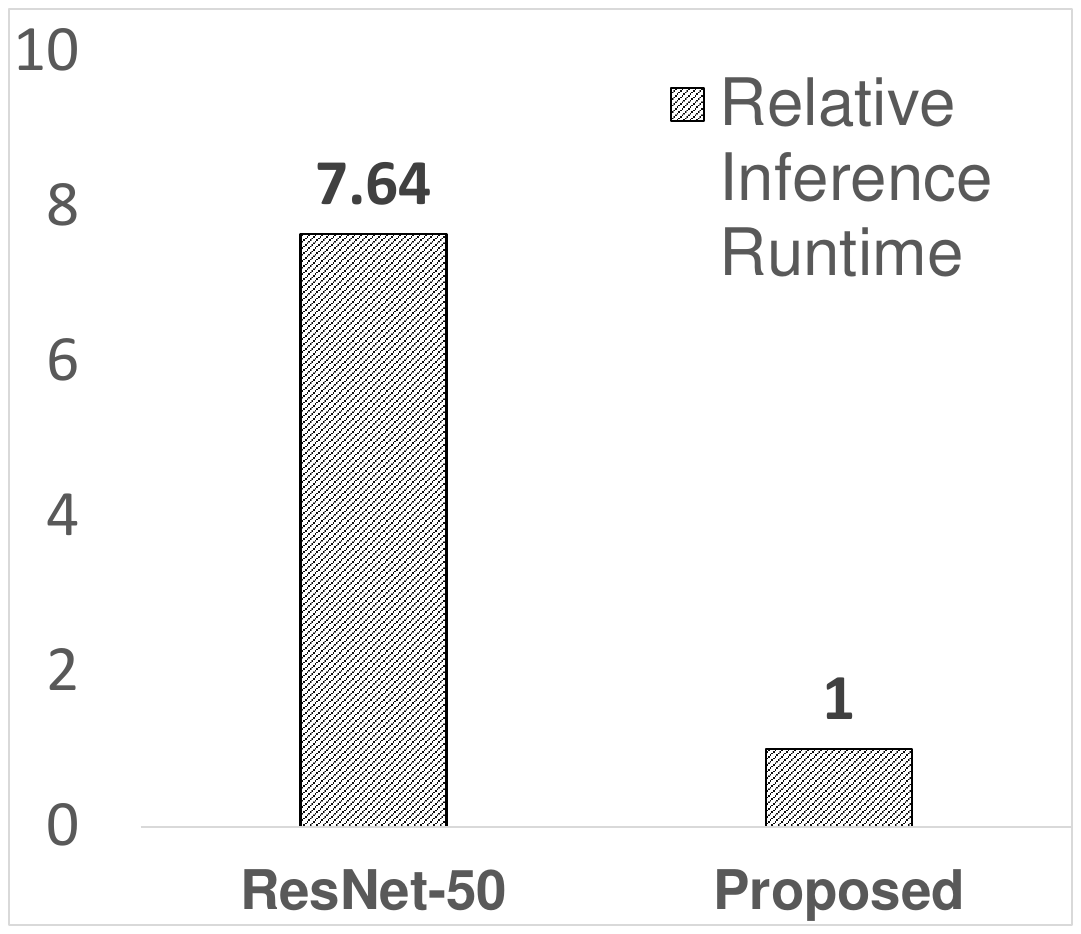}
\caption{The number of parameters (left) and the inference runtime (right) of our proposed ensemble relative to the popular ResNet-50 architecture.}
%\vspace{-0.3cm}
\label{fig:paramsAndtime}
\end{figure}

\section{Evaluation}
To evaluate our approach, we created a dataset of approximately 14,000 images, consisting of around 8,000 negative and 6,000 positive images obtained from real explosion footage frames of videos. We split the dataset into training and validation sets, with the validation set comprising 20\% of the whole dataset. Our models were implemented on an Intel X86 64bit machine running Ubuntu 14.04.5 LTS, using Keras 2.3.1 with Tensorflow 1.13.1 back-end. We trained each of our models for 400 epochs and saved the best model with the lowest validation loss error.

Our evaluations of the proposed ensemble methods were conducted against the popular ResNet-50 architecture using a set of 15 test videos of varying contexts, including popular TV series such as MacGyver, Britannia, and NCIS: Los Angeles, with an average duration of around 52 minutes and an average of 78,750 frames per video, encoded in 720p and 1080p resolutions. Human operators inspected the videos in multiple rounds to provide ground truth data with the time intervals of where explosion happened, with an average of 10.75 distinct explosion scenes recorded as ground truth for an average test video.

We compared the accuracy results of our proposed ensemble model and the ResNet-50 model used as the back-end of a state-of-the-art Faster R-CNN detection network \cite{frcn} as part of our evaluation. We measured median precision, recall, and F1 score metrics of the two models on the classification task. Since the ground truth data provided by the human operators were recorded as time intervals, we converted the detection's frames to timestamps and considered a match if a detection was within a second of the recorded ground truth time. Figure \ref{fig:evaluation} illustrates how the number of parameters and inference time of our proposed ensemble method compares with the popular ResNet-50 architecture.

\begin{figure}[!b]
\centering
%\vspace{-0.5cm}
~\\ \includegraphics[width=.9\columnwidth]{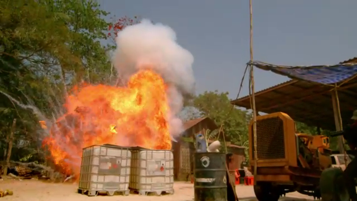}~\\
~\\ \includegraphics[width=.9\columnwidth]{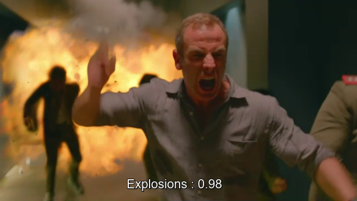}
\caption{Examples of a correctly detected explosion scenes on test videos.}
%\vspace{-0.1cm}
\label{fig:example}
\end{figure}
Figure \ref{fig:paramsAndtime} shows how the number of parameters and inference time of our proposed ensemble method compares with the popular ResNet-50 architecture. On an average video, our proposed approach was able to achieve a 100\% precision which is significantly higher than the 67\% precision made by the popular ResNet-50 model used as the back-end of a Faster R-CNN detection network. Our approach eliminated many false positives, potentially saving hundreds of hours of manual content moderation and explosion detection checks by removing the need to verify false detections in the reference videos through manual inspections. In addition, our proposed structure is significantly lighter with almost 19x fewer parameters, with the ability to decrease inference run-time by a large factor, almost 7.64x faster compared to the complex ResNet-50 model. These efficiency gains are critical in automated content moderation and explosion detection scenarios, where platforms need to moderate vast amounts of user-generated content quickly and accurately to ensure compliance with regulations and user safety. Figure \ref{fig:example} shows examples of correctly detected explosion scenes on test videos. 

\begin{figure}[!t]
\centering
%\vspace{-0.3cm}
\includegraphics[width=1\columnwidth]{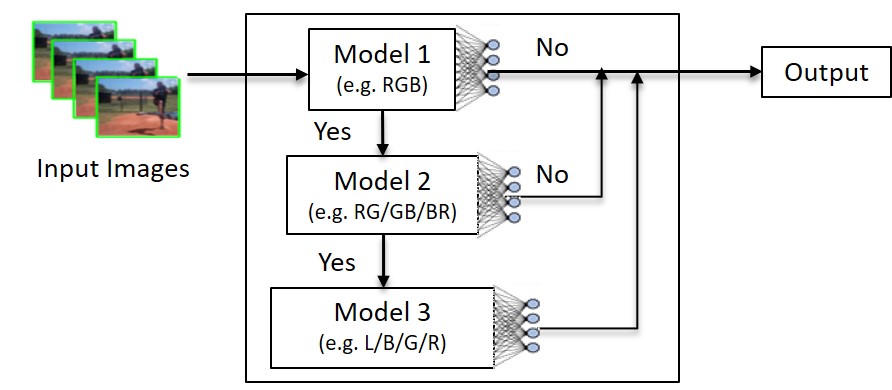}
%\vspace{-0.5cm}
\caption{An abstract extension of our proposed ensemble design.}
\label{fig:extension}
%\vspace{-0.5cm}
\end{figure}

\section{Discussion}
We believe that the design and experiments we conducted on the explosion detection use case are generalizable and can be applied to other similar image classification tasks and content moderation use cases as well. We therefore invite researchers to apply our design to other image or video classification tasks, especially those involving detection of non-rigid objects where color might be a dominant specification of the object, such as blood gore detection, smoke detection, fire detection, steam detection, and so on. The proposed ensemble approach can be useful for identifying and removing inappropriate or harmful content from online platforms or during broadcasting content moderation.

Based on the insights gained from our experiments, we propose a \textit{"think small, think many"} strategy for classification scenarios. We argue that transforming a single, large, monolithic deep model into a verification-based step-model ensemble of multiple small, simple, and lightweight models with narrowed-down features, like our shrinking color spaces, can lead to predictions with higher accuracy. In this paper, we demonstrate that our ensemble design was founded upon two base models, a primary and a secondary model, and used color spaces as the notion of features. The secondary model is fed with a limited set of the features delivered to the primary model, validating predictions made by the primary model.

We believe that our design can be extended and generalized to a validation-based ensemble of three or more base models. Figure \ref{fig:extension} depicts a sample illustration of our validation-based ensemble structure extended towards higher numbers of models. In this example extension, Model 2 validates predictions made by Model 1, and Model 3 further validates predictions made by Model 2, while the features passed get more limited as we iterate from Model 1 to Model 3. In this example, Model 1 operates on all three RGB channels, Model 2 operates on two color channels (RG, GB, or BR), and Model 3 performs validations only on a single color space, whether grayscale, or any of the R, G, or B channels. We believe this abstract concept can be generalized and extended to any feature set beyond only color spaces, which would be an avenue of exploration for the research community to consider.

In conclusion, our work demonstrates the effectiveness of ensembling multiple small, simple, and lightweight models with narrowed-down features for image classification tasks. We hope that our proposed design and the open call to apply it to other image or video classification tasks will inspire and benefit the research community.

\section{Conclusion}
\label{conclusion}
In this paper, we proposed an efficient and lightweight deep classification ensemble structure, designed for ``high-accuracy" content moderation and violence detection in videos with low false positives. Our approach is based on a set of simple visual features and utilizes a combination of lightweight models with narrowed-down color channels. We evaluated our approach on a large dataset of explosion and blast contents in TV movies and videos and demonstrated significant improvements in prediction accuracy compared to popular deep learning models such as ResNet-50, while benefiting from faster inference and lower computation cost.

Our proposed approach is not only limited to explosion detection in videos, but can be applied to other similar content moderation and violence detection use cases as well. Based on our experiments, we suggest a \textit{"think small, think many"} philosophy in deep object classification scenarios. We argue that transforming a single, large, monolithic deep model into a verification-based hierarchy of multiple small, simple, and lightweight models with narrowed-down visual features can potentially lead to predictions with higher accuracy, while maintaining efficiency and reducing computational requirements.

\bibliographystyle{IEEEtran}
\bibliography{sigproc}

\end{document}